\documentclass[runningheads]{llncs}
\usepackage[T1]{fontenc}
\usepackage{cite}
\usepackage{amsmath,amssymb,amsfonts}
\usepackage{algorithmic}
\usepackage{graphicx}
\usepackage{textcomp}
\usepackage{xcolor}
\usepackage{tabularx} 
\usepackage{lipsum}
\usepackage{multirow}
\usepackage{array}
\usepackage{booktabs}
\usepackage{makecell}
\usepackage{booktabs} 
\usepackage{fancyhdr}
\usepackage[colorlinks=true, linkcolor=black, citecolor=black, filecolor=black, urlcolor=blue]{hyperref}

\begin{document}
\pagestyle{fancy}
\fancyhf{} 
\fancyfoot[C]{\thepage} 
\renewcommand{\headrulewidth}{0pt}

\title{A Structure-aware Generative Model for Biomedical Event Extraction}
%
%
\author{Haohan Yuan\inst{1} \and
Siu Cheung Hui\inst{2} \and
Haopeng Zhang\inst{1}\thanks{corresponding author}}
\authorrunning{Y. Author et al.}
%
\institute{ALOHA Lab, University of Hawaii at Manoa,  USA \and
Nanyang Technological University, Singapore}
\maketitle              
\begin{abstract}
Biomedical Event Extraction (BEE) is a challenging task that involves modeling complex relationships between fine-grained entities in biomedical text. BEE has traditionally been formulated as a classification problem. With recent advancements in large language models (LLMs), generation-based models that cast event extraction as a sequence generation problem have attracted attention in the NLP research community. However, current generative models often overlook cross-instance information in complex event structures, such as nested and overlapping events, which constitute over 20\% of events in benchmark datasets. In this paper, we propose GenBEE, an event structure-aware generative model that captures complex event structures in biomedical text for biomedical event extraction. GenBEE constructs event prompts that distill knowledge from LLMs to incorporate both label semantics and argument dependency relationships. In addition, GenBEE generates prefixes with event structural prompts to incorporate structural features to improve the model's overall performance. We have evaluated the proposed GenBEE model on three widely used BEE benchmark datasets, namely MLEE, GE11, and PHEE. Experimental results show that GenBEE has achieved state-of-the-art performance on the MLEE and GE11 datasets, and achieved competitive results when compared to the state-of-the-art classification-based models on the PHEE dataset. Our source code is released at \url{https://github.com/HaohanYuan01/GenBEE}.

\end{abstract}
\section{Introduction}
Biomedical Event Extraction (BEE) is a challenging task that involves identifying molecular events from natural language text, which can provide invaluable information to facilitate the curation of knowledge bases and biomolecular pathways \cite{kim-etal-2009-overview}. BEE is typically divided into two subtasks, namely biomedical event trigger detection and biomedical event argument extraction. Biomedical event trigger detection aims to identify trigger words and classify them by event types that represent the presence of biomedical events. On the other hand, biomedical event argument extraction aims to identify the arguments and associate them with specific roles in biomedical events, correlating them with the corresponding event triggers.

\begin{figure}[t]
\centering
\vspace{-20pt}
\includegraphics[width=1\columnwidth,height=0.45\columnwidth]{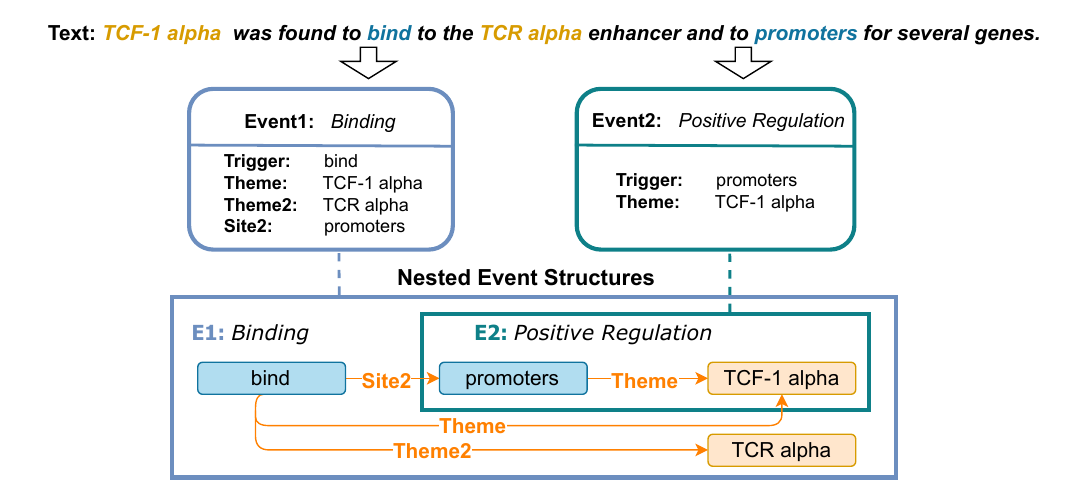}
\vspace{-15pt} 
\caption{An example of nested events identified from the given text.}
\label{NestedBEE}
\vspace{-5pt} 
\end{figure}

 Compared to flat event structures, texts in the biomedical domain often contain more nested event structures \cite{espinosa2022comparing}. For example, in benchmark datasets like MLEE \cite{pyysalo2012event} and GE11 \cite{kim-etal-2011-overview-genia}, the nested event structures constitute approximately 25\% of all events in the datasets. Figure~\ref{NestedBEE}
shows an example of two nested events identified from the given text. The trigger word \textit{``promoters''} of the \texttt{Positive Regulation} event (E2) serves as the argument role \texttt{Site} for the trigger word \textit{``bind''}, which triggers the \texttt{Binding} event (E1). Furthermore,  \textit{``TCF-1 alpha''} serves as the argument role \texttt{Theme} for both events (E1 and E2).  In addition, a trigger word may also trigger different events in a sentence, depending on its event type or arguments. Events with such structures are called overlapping events. Identifying these complex structures is important for BEE, as it can reveal the underlying relationships between events. For example, a \texttt{Site2} argument in a \texttt{Binding} event is likely to be a trigger for a \texttt{Positive Regulation} event.

\begin{figure}[t]
\centering
\vspace{-20pt}
\includegraphics[width=1\columnwidth,height=0.32\columnwidth]{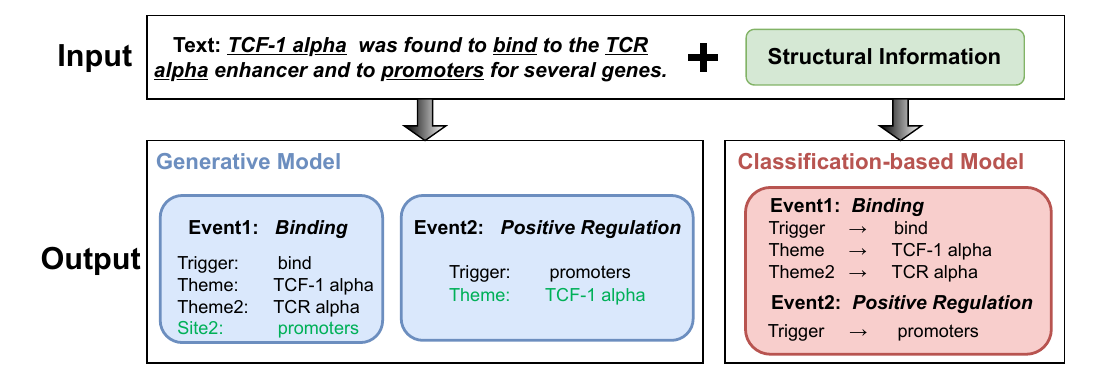}
\vspace{-15pt} 
\caption{Comparison of generative and classification-based models in extracting nested events from biomedical text.}
\label{gen-vs-cla}
\vspace{-5pt} 
\end{figure}

Traditionally, BEE has been formulated as a token classification problem \cite{bjorne2018biomedical,huang2020biomedical}. Although recent classification-based models have demonstrated a certain level of performance in BEE \cite{hao2024effective,hsu-etal-2023-tagprime}, most of them can only handle flat event structures and ignore nested events \cite{espinosa2022comparing} that appear in the given text. As shown in Figure~\ref{gen-vs-cla}, while classification-based models incorporate structural information, they fail to extract two arguments (highlighted in green) due to their inherent limitation of assigning a single class to each entity. In contrast, generative models with structural information successfully predict all triggers and arguments in the given text. Additionally, many current classification-based methods extract events in a pipeline manner \cite{wadden2019entity}, where trigger detection and argument extraction are treated as separate phases,  which hinders the learning of shared knowledge between subtasks \cite{trieu2020deepeventmine}. 

  Recently, generation-based models have been proposed for event extraction (EE) on general domains \cite{li-etal-2021-document, lu-etal-2021-text2event}. These models typically cast the event extraction task as a sequence generation problem using encoder-decoder based pre-trained language models (PLMs) \cite{raffel2020exploring, lewis-etal-2020-bart} to output conditional generation sequences.  These models often benefit from natural language prompts to learn additional semantic information  \cite{hsu-etal-2022-degree}. Apart from employing natural language prompts to enhance semantic learning from the given context, some recent works such as GTEE \cite{liu2022dynamic} and AMPERE \cite{hsu2023ampere} have also proposed injecting continuous prompts \cite{li-liang-2021-prefix,hu2021lora} into their frameworks for better contextual representation learning for event argument extraction. However, most current generation-based EE models \cite{liu2022dynamic,hsu-etal-2022-degree,hsu2023ampere} have overlooked the incorporation of the structural information. Moreover, Fei et al. \cite{NEURIPS2022_63943ee9} pointed out that the mainstay contextual representations in a generative language model's encoder tend to weaken other high-level information such as structural information, thereby reducing the effects of such information at the decoder. Furthermore, it is also unclear whether the generative language model contains sufficient domain knowledge to represent complex biomedical events. Therefore, it is a challenging task to incorporate structural information effectively into generative BEE models.

 To address the above issues, we propose an event structure-aware \textbf{Gen}erative \textbf{B}iomedical \textbf{E}vent \textbf{E}xtraction model (GenBEE), which aims to capture complex event structures effectively from biomedical texts. First, GenBEE constructs type description prompts to encapsulate the semantics of type labels. Next, it distills the knowledge from large language models (LLMs), such as GPT-4~\cite{openai2024gpt4technicalreport}, by creating event template prompts, which incorporate information such as dependencies among argument roles, to enhance contextual representation learning. Additionally, GenBEE introduces a structural prefix learning module that generates structure-aware prefixes, which are further enriched with structural prompts to provide structure-aware signals. 
 
GenBEE serves as a comprehensive BEE framework that simultaneously handles event trigger detection and event argument extraction, effectively leveraging the shared knowledge and dependencies between these two subtasks. The main contributions of this paper are summarized as follows:
\begin{itemize}
    \item We propose GenBEE, an event structure-aware generative model for biomedical event extraction. Our proposed end-to-end model can distill knowledge from the LLM to incorporate label semantics, dependency relationships among arguments, and shared knowledge between subtasks.
    \item In the GenBEE model, we propose a structural prefix learning module to guide the generation process. This module constructs structure-aware prefixes to link continuous prompts with the representation space of generative models. It is worth noting that this plug-and-play module can be incorporated into any biomedical event extraction framework. 
    \item We have evaluated GenBEE on three widely-used BEE benchmark datasets, namely MLEE, GE11, and PHEE. Experimental results show that GenBEE has achieved state-of-the-art performance on the MLEE and GE11 datasets, and achieved competitive results when compared to the state-of-the-art classification-based models on the PHEE dataset.
\end{itemize}



\section{Related Work}

\textbf{Biomedical Event Extraction.} Many current methods for biomedical event extraction adopt a ``pipeline'' strategy, where trigger detection and argument extraction are treated as separate phases. However, ``pipeline" methods may result in propagation of errors. To address this issue, Trieu et al. \cite{trieu2020deepeventmine} introduced an end-to-end method called DeepEventMine, which simultaneously identifies triggers and allocates roles to entities, thereby mitigating error propagation. However,  DeepEventMine requires the full annotation of all entities, which may not always be accessible in datasets. Ramponi et al. \cite{ramponi-etal-2020-biomedical} considered biomedical event extraction as a sequence labeling task that involves joint training models for trigger detection and argument extraction via multi-task learning. Aside from these classification-based methods, some recent works have formulated event extraction as a question-answering task \cite{wang-etal-2020-biomedical,li2020event}. This new paradigm transforms traditional classification methods into multiple questioning rounds, producing a natural language answer about a trigger or an argument in each round. Current QA-based event extraction methods are primarily focused on formulating separate questions for different events and argument types. As such, these QA methods lack the capability on capturing dependency information between arguments, which could be effective for assigning argument types.


\textbf{Generative Event Extraction.} Recently, another emerging line of work that treats event extraction as a conditional generation problem \cite{paolini2021structured, huang2021document, li2021document, hsu-etal-2022-degree} has evolved.  The development of generation-based event extraction models begins with the reformulation of event extraction problems as a generation task. For example, TANL \cite{paolini2021structured}  treated event extraction as translation tasks with augmented labels by using a variety of brackets and vertical bar symbols for representation, and tuned the model to predict these augmented labels. TempGen \cite{huang2021document}  proposed using templates for constructing role-filler tasks for entity extraction, and generating outputs that fill role entities into non-natural templated sequences. To adapt generation-based models to the task of event argument extraction, BART-Gen \cite{li2021document} investigated generative document-level event argument extraction. However, these methods only use naive natural language prompts and have not considered the incorporation of semantic information into the natural language prompts. To better incorporate label semantics into the model, DEGREE \cite{hsu-etal-2022-degree} proposed a template-based model that incorporates label semantics into the designed prompts to extract event information.  More recently, researchers have started to advance this line of work. For example,  GTEE \cite{liu2022dynamic} used prefixes \cite{li-liang-2021-prefix}, which are sequences of tunable continuous prompts, to incorporate contextual type information into the generation-based model. AMPERE \cite{hsu2023ampere}  proposed to equip the generative model with AMR-aware prefixes to embed the AMR graph, thereby providing more contextual information. To the best of our knowledge, there is no work done on integrating event structural features of complex events into generative models for BEE. Therefore, in this paper, we propose a novel generative model called GenBEE that incorporates complex event structural information for BEE.

\section{Method}
\begin{figure}[t!]
\centering
\vspace{-20pt}
\includegraphics[width=1\textwidth,height=0.4\textwidth]{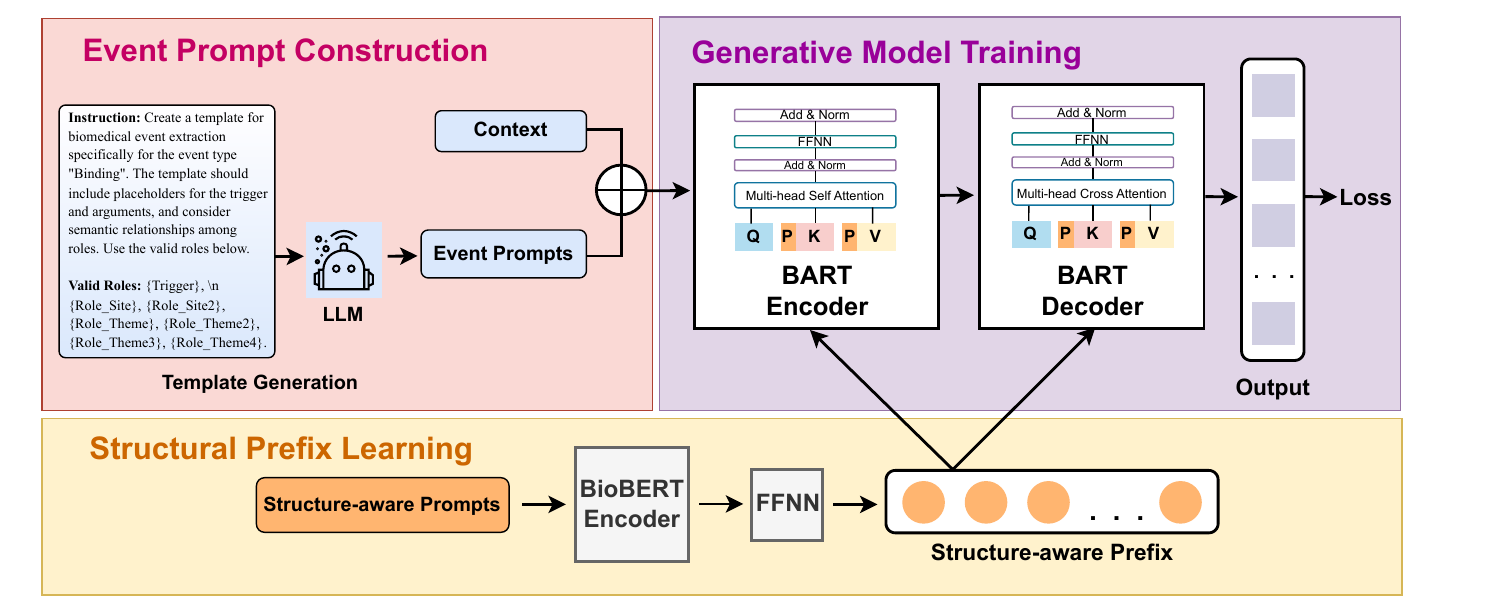}
\vspace{-15pt}
\caption{ The overall architecture of our proposed GenBEE model.}
\label{Genbee-model-architecture}
\vspace{-5pt}
\end{figure}

In this section, we discuss our proposed GenBEE model in detail. The proposed GenBEE model performs generative biomedical event extraction as follows: The extraction process is divided into a number of subtasks based on each pre-defined event type of a dataset. Given a Dataset $\mathcal{D}$ $=$ $\left\{ \mathcal{C}_j \mid j \in [1, |\mathcal{D}|] \right\}$ with its pre-defined event type set $\mathcal{E}$ $=$ $\left\{ e_i \mid i \in [1, |\mathcal{E}|] \right\}$, where $\mathcal{C}_j$ denotes a textual context and $e_i$ denotes an event type. Each subtask is denoted as $S_{e_{i},\mathcal{C}}$, for the type of event $e_i$ and the context $\mathcal{C}$, where $\mathcal{C}$ contains all the textual contexts in the dataset. As such, we have a total of $|\mathcal{E}|$ generation subtasks based on each event type for all textual contexts $\mathcal{C}$. The proposed model conducts training iteratively based on each subtask with all textual contexts in the dataset. Figure~\ref{Genbee-model-architecture} shows the architecture of the proposed model, which consists of the following modules: Event Prompt Construction, Structural Prefix Learning, and Generative Model Training.

\subsection{Event Prompt Construction}
\label{prompt design}

\begin{table}[t!]
\centering
\vspace{-20pt}
\caption{Example event prompts for the GE11 dataset.}
\label{prompts:event-types}
\renewcommand{\arraystretch}{1}
\setlength{\tabcolsep}{2pt}
\resizebox{\textwidth}{!}{ 
    \begin{tabular}{|p{2cm}|p{6cm}|p{8cm}|}
        \hline
        \textbf{Event Type} & \textbf{Type Description} & \textbf{Event Template} \\
        \hline
        Binding & Binding events involve two or more molecules coming together to form a complex. & Event trigger \{Trigger\} \textless SEP\textgreater \  \{Role\_Theme\} at binding site \{Role\_Site\} and \{Role\_Theme2\} at adjacent site \{Role\_Site2\} form a complex, assisted by \{Role\_Theme3\} and \{Role\_Theme4\}. \\
        \hline
        Positive regulation & Positive regulation events involve the activation of gene expression or signaling pathways. & Event trigger \{Trigger\} \textless SEP\textgreater \  Activator \{Role\_Cause\} at control site \{Role\_CSite\} initiates signaling at \{Role\_Site\}, enhancing the function of \{Role\_Theme\}. \\
        \hline
        Localization & Localization events track the movement of a biological entity to a specific cellular or anatomical location. & Event trigger \{Trigger\} \textless SEP\textgreater \  From \{Role\_AtLoc\}, \{Role\_Theme\} relocates to \{Role\_ToLoc\}. \\
        \hline
    \end{tabular}
}
\end{table}

 This module constructs event prompts as part of the model input to the proposed GenBEE model for capturing the semantic meaning of event types and the dependency relationships between arguments. Event prompts not only provide semantic information to the model but also define the output format. The event prompt $\mathcal{N}_{e_i}$ for event type $e_i$ contains the following components:

\begin{itemize}
    \item Event Type - It specifies the expected event type to be extracted. 
    
    \item Type Description - It provides a description on the event type.
    
    \item Event Template - It specifies the expected output format of the extracted event.  The event template consists of three parts: a trigger, a separation marker  \textless SEP\textgreater \ and a number of arguments. We use two types of placeholders, namely \textit{\{Trigger\}} and \textit{\{Role\_\textless argument\textgreater \}}, to represent the corresponding trigger and arguments, respectively.

\end{itemize}

Table~\ref{prompts:event-types} shows some example event prompts for the GE11 dataset. For example, the event type description for the \texttt{Binding} event is  ``\textit{Binding involves two or more molecules coming together to form a complex for various biological processes and signaling pathways.}". Note that the type description is obtained according to the textual description of each event type provided in the technical report of the corresponding biomedical event extraction dataset (e.g., GE11 \cite{kim-etal-2011-overview-genia}).  In the event template, the trigger part is ``\textit{Event trigger  \{Trigger\}}", which remains the same for all event types. The argument part is specific to each event type $e_i$. Each \textit{\{Role\_\textless argument\textgreater \}} serves as a placeholder for the corresponding argument role for the event.
 
 Most previous works \cite{liu2022dynamic,hsu2023ampere} proposed to use manually designed templates for event extraction. However, it is challenging and time-consuming to define event templates for each event type in the biomedical event datasets, as domain knowledge is required. In this work, we propose to elicit knowledge from Large Language Models (LLM) GPT-4 \cite{openai2024gpt4technicalreport} with designed prompts to generate the event templates for the event types pre-defined in the biomedical event datasets. Figure~\ref{Genbee-model-architecture} shows an input example that we have designed for generating the event template for the  \texttt{Binding} event using LLMs. The input prompt consists of an ``Instruction'' that specifies the generation objective for the biomedical event extraction template and ``Valid Roles'' that outline the target trigger and argument roles, which are pre-defined by the biomedical event dataset. The output of LLMs is an event template that contains richer semantic information for generative event extraction.

 After constructing the event prompt $\mathcal{N}_{e_i}$  for the event type $e_i$,  we concatenate it with the context $\mathcal{C}_j$. For each context $\mathcal{C}_j$, the input tokens $\mathcal{X}_{e_i,j}$ in the proposed model is constructed as follows: 

\begin{equation}
\mathcal{X}_{e_i,j} = [ \mathcal{N}_{e_i} ; [SEP] ; \mathcal{C}_j ]
\end{equation}
where ``$ ; $'' denotes the sequence concatenation, and $[SEP]$ is the special separator token in the Pre-trained Language Model (PLM) BART \cite{lewis-etal-2020-bart}. For each subtask based on $e_i$, we construct the input tokens for all the contexts in the dataset.

\subsection{Structural Prefix Learning}
In biomedical texts, events are often complex and interrelated.  Identifying the latent relationships among entities across different events could provide important clues for capturing complex event structures, thereby enhancing representation learning. Table~\ref{structural prompts} shows the structural prompts \cite{zhang2023instruction} constructed in the proposed model to capture different event structures for the \texttt{Binding} event. The structural prompts aim to provide descriptions of different event structures and then elicit the PLM to identify complex event structures by capturing the latent relationships among entities across different biomedical events. As shown in Table \ref{structural prompts}, the structural prompts are the same for all event types, except the corresponding event type name specified between the two tags of \textless T\textgreater. In this work, structural prompts are categorized into three groups: General Events, Overlapping Events, and Nested Events. The General Events describe the structural information of common events, guiding the PLM to identify potential interactions between frequently co-occurring events. The Overlapping Events describe the structural information of overlapping events, instructing the PLM to identify trigger words that commonly trigger overlapping events. The Nested Events describe the structural information of nested events, guiding the PLM to identify entities that frequently serve as argument roles in the events while acting as trigger words in others, or vice versa. Overall, the structural prompts are designed to describe the various structural information of biomedical events for generative event extraction.

\begin{table}[t] 
    \centering
    \vspace{-20pt}
    \renewcommand{\arraystretch}{1}
    \scriptsize 
    \caption{Example structural prompts for the \texttt{Binding} event.}
    \label{structural prompts}
    \resizebox{0.8\textwidth}{!}{ 
        \begin{tabular}{|p{1.5cm}|p{6.5cm}|} 
            \hline
            \textbf{Category} & \textbf{Structural Prompts} \\
            \hline
            General Events & (S1) What events are frequently \textbf{co-occurring} with \textless T\textgreater \texttt{Binding}\textless T\textgreater\ events? \\ 
            \hline
            Overlapping Events & (S2) What entities are frequently serving as \textbf{triggers} in both \textless T\textgreater \texttt{Binding}\textless T\textgreater\ events and other event types? \\
            \hline
            Nested Events & (S3) What entities are frequently acting in \textbf{multiple roles}, including as roles in \textless T\textgreater \texttt{Binding}\textless T\textgreater\ events and differently in other events? \\
            & (S4) What entities are frequently acting \textbf{triggers} of \textless T\textgreater \texttt{Binding}\textless T\textgreater\ events while acting as a \textbf{role} in other events, or vice versa? \\
            \hline
        \end{tabular}\vspace{-5pt}
    }
    \vspace{-10pt}
\end{table}

After structural prompts are constructed, this module generates prefixes with an encoder-only PLM to embed structural information of biomedical events from the structural prompts. In particular, BioBERT is used to generate trainable soft prompts to the BART Encoding-Decoding module of GenBEE, guiding its generation process with event structural information. 

First, structural prompts are concatenated with the separator token $[SEP]$ to construct a textual sequence $S_{e_i}$ as follows:
{\small \begin{equation}
S_{e_i} = \langle [CLS],S1_{e_i},[SEP],S2_{e_i},[SEP], ..., S4_{e_i},[SEP] \rangle
\end{equation}}where $S1_{e_i}$ to $S4_{e_i}$ are structural prompts for event type $e_i$. Then, the pre-trained BioBERT is employed to  encode the textual sequence $S_{e_i}$ for event type $e_i$ into dense vectors $Dense_{e_i}$ as follows:
\begin{equation}
Dense_{e_i} = BioBERT({S}_{e_i})
\end{equation}

After obtaining the dense vectors $Dense_{e_i}$, we extract the representation $h_{[CLS]}$ from $Dense_{e_i}$, which is encoded from the [CLS] token. The [CLS] token is the start marker of the input to BioBERT. As the first vector in $Dense_{e_i}$, $h_{[CLS]}$ contains high-level event structural information distilled from BioBERT. We then input $h_{[CLS]}$ into a  Feed Forward Neural Network (FFNN) to model the prefix $\mathcal{P}_{e_i}$ as follows:
\begin{equation}
\mathcal{P}_{e_i} = FFNN(h_{[CLS]})
\end{equation}
The length $l$ of prefix $\mathcal{P}_{e_i}$ is a hyper-parameter, which is determined experimentally, to control the size of the soft prompts.

\subsection{Generative Model Training}
The proposed GenBEE model employs a generative pre-trained language model BART for encoding and decoding. In this module, the prefix $\mathcal{P}_{e_i}$ is integrated into both the encoder and decoder of the model. Specifically, from the prefix for event type $e_i$, we obtain the additional key and value matrices as $\mathcal{P}_{e_i} = K^{e_i} = V^{e_i}$, which can further be expressed as $K^{e_i} = \{ k_{1}^{e_i},...,k_{l}^{e_i} \}$ and $V^{e_i} = \{ v_{1}^{e_i},...,v_{l}^{e_i} \}$. Here, $k_{n}^{e_i}$ and $v_{n}^{e_i}$ $(n \in [1, l])$ are vectors with the same hidden dimension in the Transformer layer. These additional key and value matrices are concatenated with the original key and value matrices in the self-attention layers of the encoder and cross-attention layers of the decoder. As such, when computing the dot-product attention, the query matrix at each position is influenced by these structure-aware prefixes, which will subsequently influence the weightings of the representations for event generation. Note that the prefix $\mathcal{P}_{e_i}$ is learnable according to the querying event type $e_i$, thereby guiding the PLM BART to differentiate event types with different structural features. 

Given the input tokens $\mathcal{X}_{e_i,j}$ and the injected prefixes $\mathcal{P}_{e_i}$,  the model first computes the hidden vector representation $H$ of the input $\mathcal{X}_{e_i,j}$ with the injected prefixes $\mathcal{P}_{e_i}$ using a bidirectional BART Encoder:
\begin{equation}
H = \text{Encoder}(\mathcal{P}_{e_i},\mathcal{X}_{e_i,j})
\end{equation}
where each layer of the BART Encoder is a Transformer block with the self-attention mechanism.

The BART Decoder then takes in the hidden states $H$ from the BART Encoder to generate the text $\mathcal{Y}_{e_i,j} = \{ y_1, \cdots, y_n \}$ ($n \in [1,|\mathcal{Y}_{e_i,j}|]$) token by token (conditioned on its previous contexts). The injected prefix $\mathcal{P}_{e_i}$ and the hidden states $h^D$ of the decoder $D$ are also involved in the computation. More specifically, during the $n$-th step of generation, the autoregressive BART Decoder predicts the $n$-th token $y_n$ and computes the $n$-th hidden state $h_n^D$ of the decoder as follows:
\begin{equation}
(y_n,h_n^D) = \text{Decoder}(y_{n-1}, [\mathcal{P}_{e_i};H; h_1^D, \ldots, h_{n-1}^D])
\end{equation}
    Each layer of the BART Decoder is a Transformer block that includes two types of attention mechanisms: a self-attention mechanism and a cross-attention mechanism. The self-attention mechanism uses the decoder's hidden state $h_n^D$ to reference the previously generated words in the decoder's output sequence.  The cross-attention mechanism uses both the encoder's hidden state $H$ and the decoder's hidden state $h_n^D$ to integrate contextual representation into the decoding process.

Then, this module outputs a structured sequence that starts with the start token `` \textless BOS\textgreater ” and ends with the end token `` \textless EOS\textgreater ”. Given the input sequence $\mathcal{X}_{e_i,j}$ with the injected prefixes $\mathcal{P}_{e_i}$,  the conditional probability $p(\mathcal{Y}_{e_i,j}|\mathcal{P}_{e_i},\mathcal{X}_{e_i,j})$ of the  output sequence $\mathcal{Y}_{e_i,j}$ is calculated progressively by the probability of each step as follows:

{\small \begin{equation}
p(\mathcal{Y}_{e_i,j}|\mathcal{P}_{e_i},\mathcal{X}_{e_i,j}) = \prod_{n=1}^{|\mathcal{Y}_{e_i,j}|} p(y_n| y_1, \cdots, y_{n-1},\mathcal{P}_{e_i},\mathcal{X}_{e_i,j})
\end{equation}}

Finally, the BART Decoder calculates the conditional probability of its output and generates the sequence $\mathcal{Y}_{e_i,j}$ for event type $e_i$ and context $\mathcal{C}_j$ as follows:
\begin{equation}
    \mathcal{Y}_{e_i,j} = \text{Decoder}(\mathcal{P}_{e_i},\mathcal{X}_{e_i,j})
\end{equation}
where $\mathcal{P}_{e_i}$ denotes the structural prefix for event type $e_i$ and $\mathcal{X}_{e_i,j}$ denotes the input tokens for event type $e_i$ with context $\mathcal{C}_j$. After obtaining the output tokens $\mathcal{Y}_{e_i,j}$, the proposed model can then perform training and inference.

For training, the trainable parameters of the GenBEE model include those in the encoder-only pre-trained language model BioBERT and the generative pre-trained language model BART. The training objective of the GenBEE model is to minimize the negative log-likelihood of the ground truth sequence $\mathcal{Y'}_{e_i,j}$, given the output sequence $\mathcal{Y}_{e_i,j}$ as follows:

\begin{equation}
Loss = -log \sum_{i=1}^{|\mathcal{E}|} P(\mathcal{Y'}_{e_i,j} | \mathcal{Y}_{e_i,j}) 
\end{equation}

        GenBEE is specifically trained to perform prediction based on the given event type. During training, it learns and incorporates the structural features of each event type. If the context $\mathcal{C}_j$ contains multiple events, the GenBEE model will generate output text for each event template, with each corresponding to a trigger and its associated argument roles.  If the model does not give any prediction result on any triggers or argument roles for a given event type, the output will contain the corresponding placeholders only.

For inference, the proposed GenBEE model enumerates all event types and generates outputs for each event type based on the given context. After generating the output, the model compares the generated tokens with the specified event templates for each event type to identify the triggers and arguments accordingly. Finally, string matching \cite{hsu-etal-2022-degree} is employed to identify and capture the span offsets of the predicted triggers and arguments. 

\section{Experiments}
In this section, we first describe the dataset, evaluation metrics, implementation details, and baseline models. Then, we present the experimental results to show the effectiveness of the proposed GenBEE model for biomedical event extraction.

\begin{table}[t]
\centering
\vspace{-20pt}
\renewcommand\arraystretch{1}
\caption{Statistics of the datasets.}\label{dataset}
\fontsize{6}{5}\selectfont
\resizebox{1\textwidth}{!}{
\setlength\tabcolsep{5pt} 
\begin{tabular}{lccccccccc}
  \toprule
  & \multicolumn{3}{c}{MLEE} & \multicolumn{3}{c}{GE11} & \multicolumn{3}{c}{PHEE} \\
  \cmidrule(r){2-4} \cmidrule(lr){5-7} \cmidrule(l){8-10}
  & Train & Dev & Test & Train & Dev & Test & Train & Dev & Test \\
  \midrule
  \# Documents & 131 & 44 & 87 & 908 & 259 & - & - & - & - \\
  \# Sentences & 1294 & 467 & 885 & 7926 & 2483 & - & 2897 & 965 & 965 \\
  \# Events & 3121 & 670 & 1894 & 10310 & 3250 & - & 3003 & 1011 & 1005 \\
  \# Nested \& Overlapping Events & 773 & 397 & 315 & 2843 & 658 & - & 69 & 26 & 29 \\
  \# Arguments & 2887 & 1065 & 1887 & 6823 & 1533 & - & 15482 & 5123 & 5155 \\
  \bottomrule
\end{tabular}
}
\vspace{-10pt}
\end{table}

\subsection{Experimental Setup}
\textbf{Datasets.} We have conducted experiments on the following three publicly available benchmark datasets, namely MLEE, GE11, and PHEE, for biomedical event extraction. MLEE \cite{pyysalo2012event}  is obtained from the BioNLP-09 Shared Task, which concentrates on recognizing molecular events in biomedical texts. It contains 29 event types and 14 argument roles. We use the train/dev/test split given by the data provider. GE11 \cite{kim-etal-2011-overview-genia} is sourced from the BioNLP 2011 Shared Task. This dataset focuses on events associated with transcription factors in the domain of human hematopoietic cells. It contains 9 event types and 10 role arguments. We use the train/dev/test split given by the shared task and evaluate the performance based on the development set, as the test set is unannotated and the official tool for evaluation is no longer available. The PHEE dataset \cite{sun2022phee} is designed for pharmacovigilance and facilitates a comprehensive analysis of the effects of medical treatments. It contains 2 event types and 16 role arguments, with few nested event structures. We use this dataset to assess the performance of our proposed model on flatter datasets.  We use the train/dev/test split provided by TextEE \cite{huang-etal-2024-textee} and preprocess each dataset accordingly. Table~\ref{dataset} presents the statistics of the datasets.  ~\\

\begin{table*}[!t]
\centering
\renewcommand\arraystretch{1}
\caption{\label{Genbee-main-result} Experimental results (\%) for extracting events based on the MLEE, GE11 and PHEE datasets. The best score is highlighted in bold, and the second-best score is underlined.}
\fontsize{7}{4}\selectfont 
\resizebox{1\textwidth}{!}{ 
\setlength\tabcolsep{5pt} 
\begin{tabular}{l l c c c c c c c}
\toprule
  \multirow{2}{*}{Methods} &\multirow{2}{*}{Type}&\multirow{2}{*}{PLM}& \multicolumn{2}{c}{MLEE} & \multicolumn{2}{c}{GE11} & \multicolumn{2}{c}{PHEE} \\
   \cmidrule{4-5}\cmidrule{6-7}\cmidrule{8-9}
   & & & Trg-C & Arg-C & Trg-C & Arg-C & Trg-C & Arg-C\\
\midrule
DyGIE++ (2019) & Cls & RoBERTa-l & 80.6 & 65.8 & 67.1 & 62.8 & 70.1 & \textbf{53.9} \\
OneIE (2020) & Cls & RoBERTa-l & \underline{80.9} & 65.2 & 67.3 & \underline{63.9} & 69.8 & 52.0 \\
EEQA (2020) & Cls & RoBERTa-l & 79.3 & 65.3 & 66.6 & 62.7 & \underline{70.3} & 53.1 \\
AMR-IE (2021) & Cls & RoBERTa-l & 80.2 & 66.4 & 67.9 & 63.3 & 69.7 & 53.5 \\
TagPrime (2023) & Cls & RoBERTa-l & 80.6 & \underline{67.1} & \textbf{68.4} & 63.6 & \textbf{70.9} & 52.2 \\

DEGREE (2022)  & Gen & BART-l & 78.0 & 64.6 & 65.2 & 60.5 & 67.6 & 51.4 \\
\textbf{GenBEE} & Gen & BART-l & \textbf{81.4} & \textbf{67.9} & \underline{68.2} & \textbf{64.4} & 69.8 & \underline{53.8} \\
\bottomrule
\end{tabular}
}
\vspace{-10pt}
\end{table*}

\noindent\textbf{Metrics.} For evaluation metrics, we follow the previous work \cite{wadden2019entity,lu-etal-2021-text2event,hsu-etal-2022-degree} to adopt the following two metrics to evaluate the performance: (1)  Trigger Classification (Trg-C) F1-score: A trigger is correctly classified if the predicted span and the predicted event type match the gold ones.  (2) Argument Classification (Arg-C) F1 score: An argument is correctly classified if the predicted span, event type, and role type match the gold ones.  ~\\

\noindent\textbf{Implementation Details.} For a fair comparison with recent works, we leverage the BART-large \cite{lewis-etal-2020-bart} as the PLM in our proposed GenBEE model. For encoding structural prompts, we use a pre-trained BioBERT \cite{liu2019roberta} model. We train GenBEE on an NVIDIA A100 40G GPU.  The learning rate of BART-large is set to $10^{-5}$ and the learning rate of BioBERT is set to $10^{-6}$.  Moreover, we train GenBEE for 50 epochs on PHEE and 80 epochs on MLEE and GE11. The batch size is set to 16 during training. The prefix length $l$ is set to 40, which is determined experimentally within the interval  $l$ = \{20, 30, 40, 50, 60\}. ~\\

\noindent\textbf{Baseline Models.} We compare our GenBEE with the following EE models: (1) DyGIE++ \cite{wadden2019entity}  is a classification-based model that captures contextual information with the span graph propagation technique. (2) OneIE \cite{lin2020joint} is a classification-based model that utilizes global feature-aware graphs to capture cross-subtask and cross-instance interactions.  (3) AMR-IE \cite{zhang-ji-2021-abstract} is a classification-based model that captures syntactic characteristics in contexts using the AMR graph. (4) EEQA \cite{du-cardie-2020-event} is a classification-based model that formulates EE as a question-answering task.
(5) TagPrime \cite{hsu-etal-2023-tagprime} is a state-of-the-art classification-based EE model that iteratively extracts event information based on each event type. (6) DEGREE \cite{hsu-etal-2022-degree} is a state-of-the-art generation-based EE model that integrates prompts for conditional generation. As the vanilla DEGREE model was designed mainly for general domain event extraction, we have re-implemented it using the prompts provided by \cite{huang-etal-2024-textee} for the MLEE, GE11, and PHEE datasets. To ensure a fair comparison across models, we
adopt the official codes of the above baselines and train them with the same data.

\subsection{Experimental Results}

Table~\ref{Genbee-main-result} shows the performance results based on the MLEE, GE11, and PHEE datasets. ``Gen" and ``Cls" denote Generation-based models and Classification-based models, respectively. The letter ``l'' in the column PLM is used to denote the large model. Overall, the GenBEE model achieves new state-of-the-art performance on the MLEE and GE11 datasets. Moreover, on the PHEE dataset, the GenBEE model achieves the second-highest Arg-C F1-score among all baselines, while its Trg-C F1-score is still competitive with other classification-based models.  Additionally, we observe that on the PHEE dataset, which only contains few nested and overlapping event structures, GenBEE exhibits a marginal reduction in performance.

Table~\ref{Genbee-main-result} also shows that GenBEE significantly outperforms the state-of-the-art generation-based model, DEGREE, in the experiments based on the MLEE, GE11, and PHEE datasets.
 More specifically, GenBEE demonstrates an improvement of +3.4\%, +3.0\%, and +2.2\% on Trg-C F1-score, and an improvement of +3.3\%, +3.9\% and +2.4\% on Arg-C F1-score, over DEGREE on the MLEE, GE11, and PHEE datasets, respectively.  We attribute GenBEE's performance improvement to the use of event prompts and structural prefixes, which provide event structural information for biomedical event extraction.

For the experiments on the PHEE dataset, it is worth noting that the two state-of-the-art models, TagPrime and DyGIE++, are both sequence tagging models. We find that these two models mainly benefit from better span identification of trigger and argument words for achieving better performance. Even though the Trg-C F1-score of GenBEE slightly lags behind some baselines, GenBEE still achieves the second-highest  Arg-C F1-score when compared with all the baselines. We attribute this to GenBEE's end-to-end extraction style, which can be less susceptible to the error propagation issue that commonly affects pipeline methods. \vspace{-5pt}

\begin{table}[!t]
\centering
\vspace{-20pt}
\renewcommand\arraystretch{1.1}
\caption{\label{ablation-study} Ablation study on the proposed GenBEE model based on different model configurations.  We report the results in F1-scores (\%).  }
\fontsize{7}{6}\selectfont 
\resizebox{1\textwidth}{!}{ 
\setlength\tabcolsep{4pt} 
\begin{tabular}{l c c c c c c c c}
\toprule%
  \multirow{2}{*}{Model} & \multicolumn{2}{c}{MLEE} & \multicolumn{2}{c}{GE11} & \multicolumn{2}{c}{PHEE} & \multicolumn{2}{c}{$\triangle$ Average} \\
   \cmidrule{2-3}\cmidrule{4-5}\cmidrule{6-7}\cmidrule{8-9}%
  &  Trg-C  & Arg-C & Trg-C  &  Arg-C & Trg-C  &  Arg-C & Trg-C  &  Arg-C\\
\midrule
GenBEE & 81.4 & 67.9 & 68.2 & 64.4 & 69.8 & 53.8 & - & - \\
(1)  w/o Event Prompt Construction        & 80.9 & 67.3 & 67.5 & 63.0 & 69.2 & 52.9 &  -0.6 & -1.0  \\
(2)  w/o Structural Prefix Learning               & 79.8 & 66.7 & 66.4 & 62.1 & 68.7 & 52.6 & -1.5 & -1.6 \\
(3)  w/o (1) and (2)               & 78.0 & 64.6 & 65.2 & 60.5 & 67.6 & 51.4 & -2.9  & -3.2 \\
\bottomrule
\end{tabular}
}
\end{table}

\subsection{Ablation Study}
We conduct an ablation study of the proposed GenBEE model to evaluate the effects of different modules on the overall performance across the MLEE, GE11, and PHEE benchmark datasets. Specifically, we focus on the following two modules: Event Prompt Construction and Structural Prefix Learning. Table~\ref{ablation-study} reports the performance results of the ablation study. The symbol $\triangle$ indicates the difference in the F1-score between the different configuration models and the proposed GenBEE model. As shown in Table~\ref{ablation-study}, we observe that event prompts and structural prefixes significantly contribute to the performance of our proposed model, regardless of the dataset. Removing the Event Prompt Construction module leads to a decrease of -0.6\% and -1.0\% on the average Trg-C and Arg-C F1-scores, respectively. This drop in performance indicates that the prompts constructed for describing events significantly contribute to the improvement of the proposed GenBEE model. Moreover, the removal of the Structural Prefix Learning module from GenBEE reduces the performance by -1.5\% and -1.6\% on the average Trg-C and Arg-C F1-scores, respectively, highlighting the important role of this module in the model’s overall performance. Finally, removing both Event Prompt Construction and Structural Prefix Learning modules will further degrade the model's performance. Overall, each module of the GenBEE model plays a crucial role in achieving promising performance for biomedical event extraction. \vspace{-5pt}

 \begin{figure}[t]
\centering
\vspace{-20pt}
\includegraphics[width=0.8\textwidth,height=0.37\textwidth]{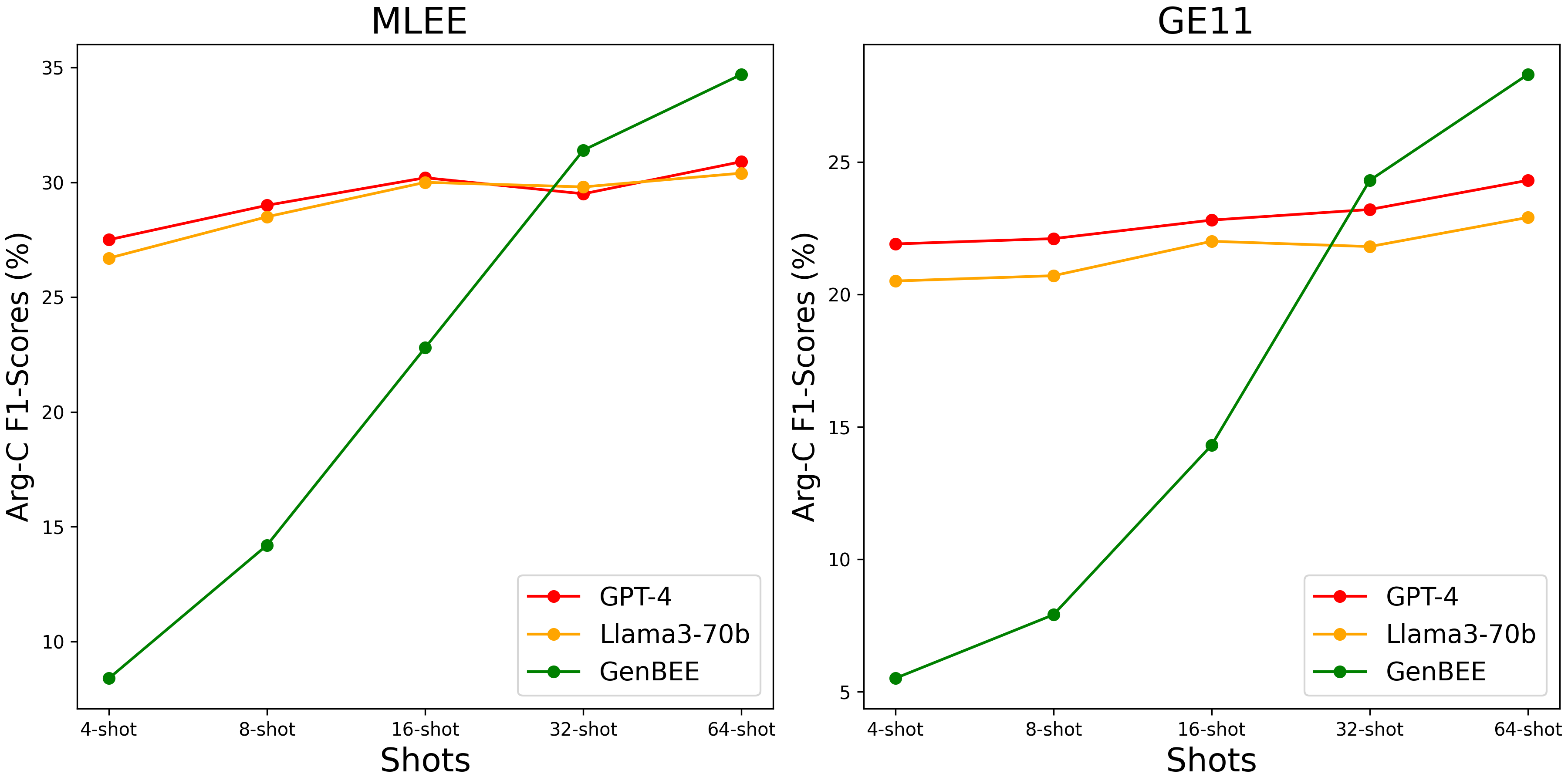}
\vspace{-10pt}
\caption{\label{comparison-LLMs} Experimental results using LLMs  and GenBEE with few-shot learning based on the MLEE and GE11 datasets. All reported results are in Arg-C F1-scores (\%).}
\label{case-study}
\vspace{-5pt}
\end{figure}
\vspace{-5pt}

\subsection{Few-shot Learning Performance Comparison with LLMs}

Given the immense potential of using large language models (LLMs) API with In-Context Learning (ICL) \cite{brown2020language} across various NLP tasks under data efficient scenarios \cite{zhang2024stancedetectiontechniquesevolve}, we conduct experiments to compare the few-shot performance of the proposed  GenBEE model with LLMs. Specifically, we consider two widely used LLMs: GPT-4 \cite{achiam2023gpt} and Llama3-70b \cite{touvron2023llama} for comparison. We access these LLMs through APIs from their official providers. As a part of the prompt,  we provide LLMs with the same type of instructions, EE templates that we use in GenBEE, and a few ICL  examples (positive ones). It is worth noting that the number of ICL examples is restricted by the maximum context length supported by LLMs.

Figure~\ref{comparison-LLMs} shows the few-shot learning performance results for LLMs and our proposed GenBEE model across the MLEE and GE11 benchmarks. The results indicate that while GPT-4 and Llama3-70b demonstrate significant advantages in extremely data-scarce scenarios, such as 4-shot and 8-shot settings, GenBEE begins to close the performance gap as the number of labeled examples increases, and eventually outperforms the LLMs. For example, in the 32-shot and 64-shot settings, the GenBEE model surpasses both LLMs. Generally, while LLMs with In-Context Learning are well-suited to extremely data-scarce scenarios, the fine-tuned GenBEE is able to outperform LLMs as the amount of labeled data increases.

 \begin{figure}[!t]
\centering
\vspace{-5pt}
\includegraphics[width=0.9\textwidth,height=0.35\textwidth]{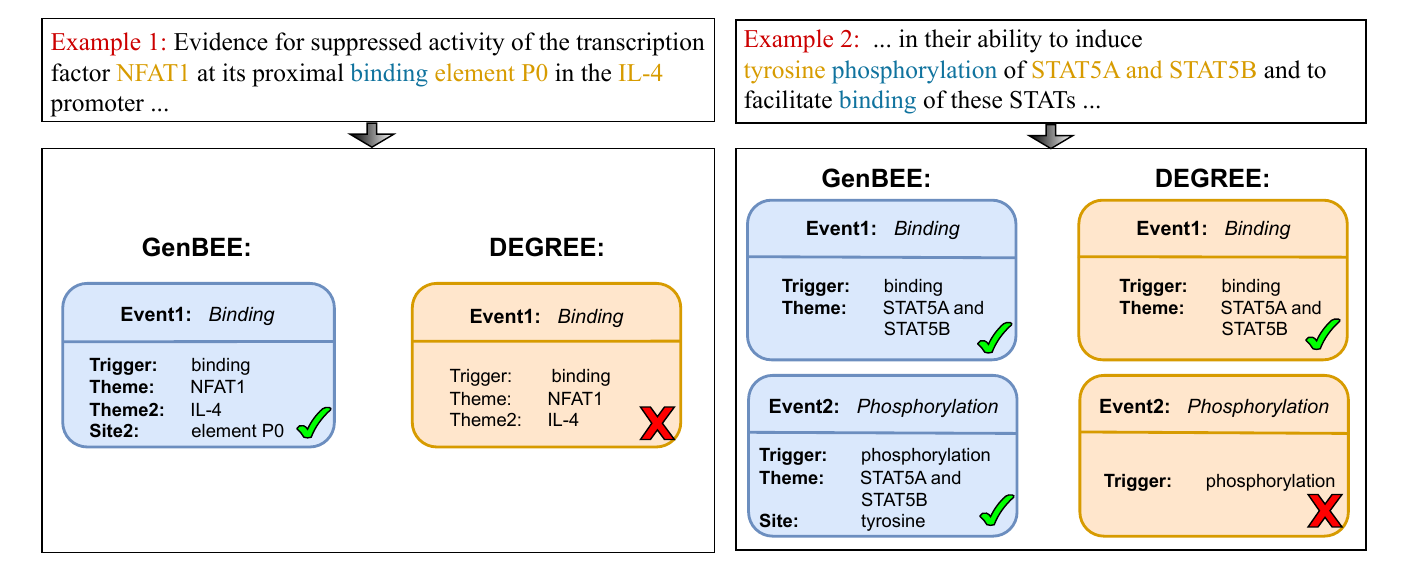}
\vspace{-5pt}
\caption{ A case study based on two examples taken from the GE11 dataset.  }
\label{case-study}
\vspace{-5pt}
\end{figure}

\vspace{-5pt}
\subsection{Case Study}

In this section, we compare GenBEE and DEGREE based on two examples, which are presented in Figure~\ref{case-study}, to illustrate the differences in their predictions. Example 1 presents a case where the  DEGREE model incorrectly predicts the argument \texttt{Site2} in the \texttt{Binding} event triggered by ``\textit{binding}'', but our GenBEE model gives the correct prediction. We infer that our model has incorporated dependency information between arguments using event prompts for enhancing the performance of biomedical event extraction. Example 2 presents a case where the DEGREE model fails to predict the arguments \texttt{Theme} and \texttt{Site} in the \texttt{Phosphorylation} event, which is nested with the \texttt{Binding} event. However, the GenBEE model is able to predict the arguments of these two nested events correctly. We infer that our model has incorporated event structural information using structural prefixes for effectively recognizing the relationships between nested events. As illustrated by the case study, with event prompts and structural prefixes, our proposed GenBEE model is able to perform biomedical event extraction effectively.

\section{Conclusion}
In this paper, we propose a novel event structure-aware generative model GenBEE, which can effectively tackle complex event structures such as nested events and overlapping events, for biomedical event extraction. The experimental results demonstrate that GenBEE outperforms strong baselines on the MLEE, GE11, and PHEE datasets. In addition, our experiments also show that structure-aware prefixes can effectively serve as a medium to link structural prompts with the representation space of generative models, thereby enhancing the model’s overall performance. 

\section*{Acknowledgment}

This work was enabled in part by funding from the National Science Foundation awards: 2201428, 2232862, 2004014, 2138259, and 2406251. We appreciate valuable suggestions from
the anonymous reviewers.
%
%
%
 \bibliographystyle{splncs04}
%
\bibliography{mybibfile}%

\end{document}